\documentclass[10pt,twocolumn,letterpaper]{article}

\usepackage[pagenumbers]{cvpr} 

\usepackage{lipsum}
\usepackage{multirow}
\usepackage{multicol}
\usepackage{scontents}
\usepackage[normalem]{ulem}
\usepackage{balance}

\usepackage[dvipsnames]{xcolor}

\renewcommand{\paragraph}[1]{\vspace{.5em}\noindent\textbf{#1}.}

\newcommand{\eq}[1]{(\ref{eq:#1})}

\usepackage[OT2,T1]{fontenc}
\DeclareSymbolFont{cyrletters}{OT2}{wncyr}{m}{n}
\DeclareMathSymbol{\Sha}{\mathalpha}{cyrletters}{"58}
\DeclareMathOperator*{\argmin}{arg\,min}

\newcommand{\x}{\mathbf{x}}

\newcommand{\freq}{\boldsymbol{\omega}}
\newcommand{\params}{{\boldsymbol{\theta}}}

\newcommand{\field}{f}
\newcommand{\f}{\field}
\newcommand{\kernel}{\kappa}

\newcommand{\expectation}{\mathbb{E}}

\newcommand{\view}{\mathbf{v}}
\newcommand{\sdf}{s}

\definecolor{cvprblue}{rgb}{0.21,0.49,0.74}
\usepackage[pagebackref,breaklinks,colorlinks,citecolor=cvprblue]{hyperref}

\def\paperID{4179} 
\def\confName{CVPR}
\def\confYear{2024}

\title{BANF: Band-limited Neural Fields for Levels of Detail Reconstruction}

\author{
  Akhmedkhan (Ahan) Shabanov\textsuperscript{1},
  Shrisudhan Govindarajan\textsuperscript{1},
  Cody Reading\textsuperscript{1}, 
  Lily Goli\textsuperscript{2}, 
  Daniel Rebain\textsuperscript{3},
  \\
  Kwang Moo Yi\textsuperscript{3},
  Andrea Tagliasacchi\textsuperscript{1, 2 , 4}
  \\[1em]
  \textsuperscript{1}Simon Fraser University,
  \textsuperscript{2}University of Toronto
  \textsuperscript{3}University of British Columbia, 
  \textsuperscript{4}Google DeepMind\\
}

\begin{document}
\twocolumn[{%
\renewcommand\twocolumn[1][]{#1}%
\maketitle
\vspace{-1em}
\includegraphics[width=\linewidth]{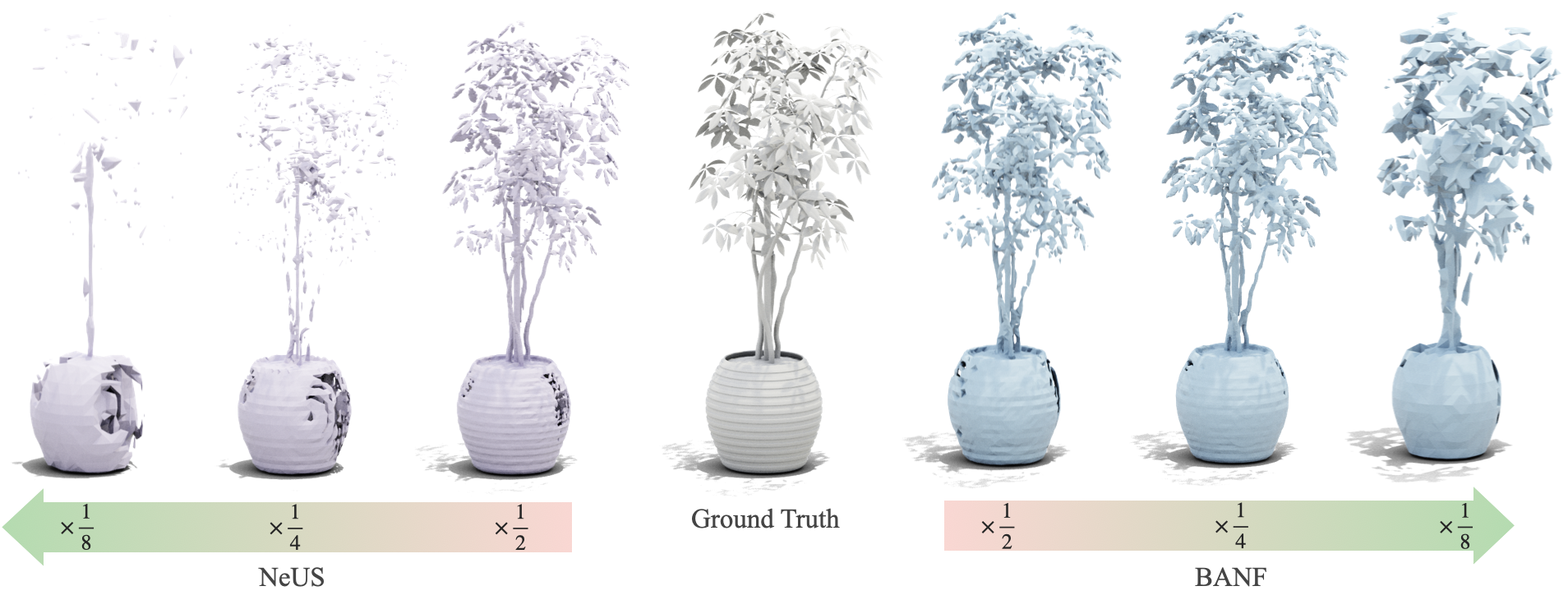}
\vspace{-1.5em}
\captionof{figure}{We introduce BANF, a method for band-limited frequency decomposition in neural fields. Our minimal yet impactful enhancements to the training process achieves anti-aliased coarse-to-fine signal reconstruction, seamlessly integrating with commonly used mesh extraction techniques~\cite{MarchingCubes}, significantly surpassing prior methods in quality.
The reconstruction artifacts due to view-dependent effects~(i.e. vase bowl not being smooth) are due to the reconstruction baseline we employ, and is orthogonal to our method.
\vspace{1em}}
\label{fig:teaser}
}]

\begin{abstract}
Largely due to their implicit nature, neural fields lack a direct mechanism for filtering, as Fourier analysis from discrete signal processing is not directly applicable to these representations.
Effective filtering of neural fields is critical to enable level-of-detail processing in downstream applications, and support operations that involve sampling the field on regular grids (e.g. marching cubes).
Existing methods that attempt to decompose neural fields in the frequency domain either resort to heuristics or require extensive modifications to the neural field architecture.
We show that via a simple modification, one can obtain neural fields that are low-pass filtered, and in turn show how this can be exploited to obtain a frequency decomposition of the entire signal.
We demonstrate the validity of our technique by investigating level-of-detail reconstruction, and showing how coarser representations can be computed effectively. Source code is available at \url{https://theialab.github.io/banf}
\vfill
\end{abstract}
\section{Introduction}
\label{sec:intro}
%
Recent research enabled neural rendering to accurately model 3D scenes from multi-view images alone.
Many neural rendering techniques employ \textit{neural fields} as the underlying 3D representation  -- coordinate neural networks that can represent a variety of signals including images~\cite{martel2021acorn, sitzmann2019siren}, surface manifolds~\cite{park2019deepsdf, wang2021neus, volsdf}, and volumetric densities~\cite{mildenhall2020nerf}.
These representations are now used in a plethora of applications that include robotics~\cite{adamkiewicz2022vision}, semantic understanding~\cite{pnf}, scene capture~\cite{neurangelo}, editing~\cite{instructnerf}, and generative modeling from natural language~\cite{dreamfusion}.

While neural fields are most commonly trained to represent the \textit{entire} spectrum of signals, 
it is an established practice in classical signal processing to decompose the signal into separate frequency bands due to the many advantages it brings~\cite{pyramid}.
For example, signal decomposition allows 3D objects to be represented at multiple levels of detail, which finds application in reducing the computational cost of rendering or physics-based simulations~\cite{LOD}.
In classical signal processing, frequency decomposition can be studied via the Fourier transform~\cite{SignalProcessing}.
Signals can be decomposed through band-pass filtering, and later recombined through linear superposition.
One could apply these classical techniques by first \textit{sampling} neural fields on a regular lattice, but the Nyquist theorem tells us that signals ought to be properly filtered before being sampled so to avoid aliasing~\cite{SignalProcessing}.
This is particularly critical for methods that target surface reconstruction~\cite{neurangelo}, as these techniques often rely on marching cubes to extract manifolds from fields~\cite{MarchingCubes}, and this process involves sampling neural fields on \textit{regular} grids; see~\cref{fig:teaser} for an example.

While monte-carlo filtering of neural fields would make anti-aliased sampling possible~\cite{deng2019neural}, a critical question is whether filtering could be realized \textit{directly} at training time to minimize its impact on inference, rather than \textit{a-posteriori} as distillation~\cite{goli2023nerf2nerf}.
This can be achieved when neural fields are implemented as fully-connected networks~(MLPs).
One can build over the theory of multiplicative filter networks~\cite{fathony2021multiplicative} to realize signal decomposition for neural fields~\cite{lindell2021bacon, Polynomial, dou2023mfld}.
However, due to \textit{significant} gains in training performance, most recent methods do not implement neural fields as large MLP networks, but rather via hybrid models combining interpolated feature grids with small MLPs mapping features to the field co-domain -- this renders the theory of multiplicative filter networks to hybrid neural fields not applicable.
One exception is \citet{wu2023nffb}, but while they introduce a pyramid-like architecture, their main outcome is a compressed representation.

Interestingly, early research in neural fields exploited the duality of time and frequency scaling to realize (explicit) multi-scale training architectures~\cite{takikawa2021nglod, liu2020neural}.
\footnote{Interestingly, as we will see later, bi-linear interpolation does execute a form of filtering, but this filtering affects the field feature space, rather than, as we desire, in the field co-domain.}
However, these methods require the multi-resolution structure of the signal to either be specified \textit{a-priori}~\cite{takikawa2021nglod}, or by monitoring the training process via heuristics~\cite{liu2020neural}.
Conversely, our technique can be applied to any neural field, without any modification to the underlying architecture, and without any assumption about the training data.

Our core insight is that regularly \textit{sampling} a field, and then \textit{interpolating} this field with a band-limited kernel can be seen as a low-pass filtering operation.
We can then build band-pass signals by compositing these filters, which allows us to derive a suitable coarse-to-fine hierarchical training scheme.
%
We demonstrate the validity of our method across domains (1D, 2D, 3D), and most importantly on 3D representations trained from 2D observations.
We show how filtered representations can be extracted at any of the intermediate scales at a fidelity that is superior to that of reasonable baselines.
Further, these representations can be aggregated, matching the performance of representations that were directly trained at high-resolution only.
\section{Related work}
\label{sec:related-work}
\begin{figure*}
\begin{center}
\includegraphics[width=\textwidth]{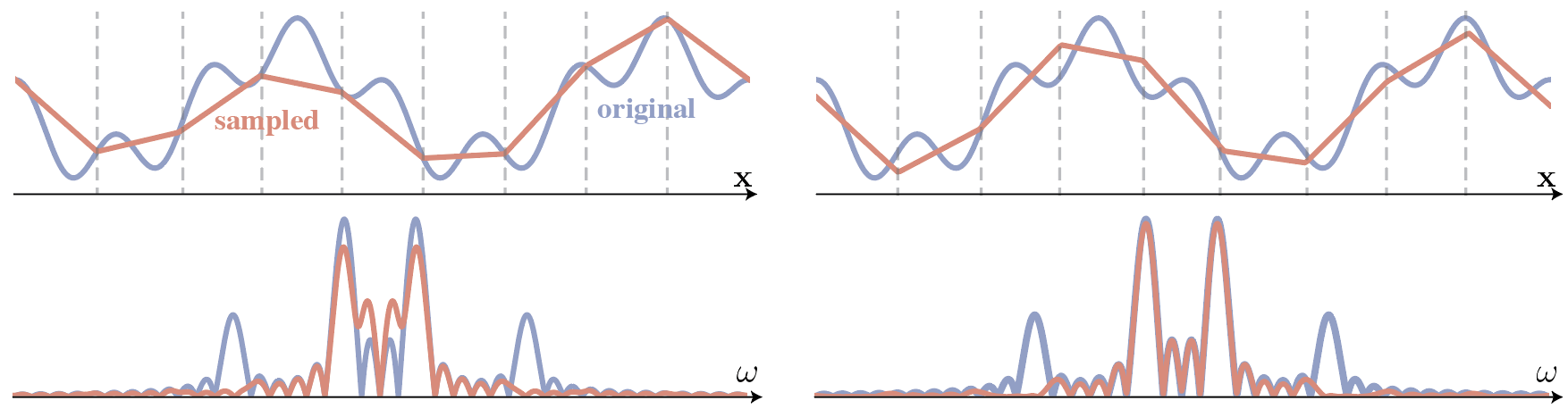}
\end{center}
\vspace{-1em}
\caption{
\textbf{Filtering by optimization} -- 
We compare two approaches to generate a low-pass version of a neural field's signal. Left shows training a neural field to a 1D signal, and sampling the result a-posteriori on a regular lattice.
On the right, our approach trains the neural field in a way that is ``sampling aware'', as the field is expressed as the~(linear) interpolation of sampled field values.
In other words, our method executes low-pass filtering \textit{during} optimization.
At the bottom we visualize the signal's spectra, where the baseline's reconstruction is clearly affected by aliasing, while our reconstruction is clearly low-pass filtered.
\vspace{-1.5em}
}
\label{fig:1d}
\end{figure*}
Our work is built on techniques in signal processing and neural fields. 
We further review how signals in neural fields can be spatially decomposed, frequency decomposed by positional encoding, and frequency decomposed by network design.

\paragraph{Frequency decomposition by signal processing} 
In traditional signal processing, signals are often converted to the frequency domain with the Fourier Transform~\cite{SignalProcessing}. 
In the frequency domain, lower frequencies represent smoother portions of the signal while high frequencies represent finer details. 
The frequency domain allows the signal to be separated into sub-bands to capture different levels of detail~(LODs)~\cite{SignalProcessing, jain1989fundamentals}. 
Unfortunately, there is no analytical Fourier Transform equivalent for signals represented as neural fields. 
A signal can be decomposed into coarse-to-fine LODs by convolution with appropriate kernels.
However, since neural fields are continuous functions, convolutions must be executed as a monte-carlo integration~\cite{deng2019neural} whose result can be distilled in a secondary neural field~\cite{goli2023nerf2nerf}. 
Conversely, we propose a simple method to directly generate low-pass and band-pass signals during field optimization, with a minimal modification in the training procedure.

\paragraph{Neural fields} As a form of signal representation, neural fields have recently gained substantial importance~\cite{NeuralFieldsVisualComputing2022}.
Neural fields can effectively represent many signals, including 2D images~\cite{martel2021acorn, sitzmann2019siren}, and 3D geometry in the form of Signed Distance Functions~\cite{park2019deepsdf} and Occupancy Fields~\cite{mescheder2019occupancy}. 
Arguably the most well-known example of neural fields is neural radiance fields,  enabling novel view synthesis through volumetric~\cite{mildenhall2020nerf} or surface~\cite{wang2021neus} rendering.
Classical neural fields~\cite{park2019deepsdf, mildenhall2020nerf} use a multilayer perceptron (MLP) to map an input coordinate to a signal value. 
Non-neural variants that optimize explicit feature grids later emerged~\cite{karnewar2022relufields, yu2022plenoxels}, sacrificing slightly worse reconstructions for faster training speed. 
Hybrid models combining explicit representations (grid~\cite{Sun2022DVGo}, hash table \cite{mueller2022instant}, K-Planes~\cite{Chen2022TensoRF}) with a small MLP finally emerged, achieving both fast training speed and high quality reconstruction. 
Although there are many variants of neural field architectures, our sampling-based strategy can be applied on top of \textit{any} architecture.

\paragraph{Multi-scale fields by spatial decomposition} Neural fields can be trained with spatial decomposition to produce coarse-to-fine LODs, where varying levels of detail are modeled with multi-resolution grids such as octrees~\cite{takikawa2021nglod} or multi-resolution tri-planes~\cite{zhuang2023lod}.
Adaptive data structures allow for increased resolution in high-frequency regions, which can be identified for further subdivision using indicators such as high local training error~\cite{martel2021acorn} or high density~\cite{liu2020neural}. Saragadam et al. \cite{saragadam2022miner} suggest a memory-efficient approach using separate MLPs for data patches when multi-scale supervision available.
Such methods are dependent on \textit{complex} and \textit{heuristic-based} spatial subdivision strategies. 
Moreover, spatial decomposition methods neglect their effect on the frequency domain, resulting in LODs that are not explicitly frequency band-limited. 
Our method does not suffer from these limitations, as we provide a \textit{simple} training scheme directly providing a frequency decomposition.

\paragraph{Frequency decomposition by positional encoding} 
\citet{tancik2020fourfeat} showed that mapping coordinates to Fourier features overcomes spectral bias in MLPs \cite{rahaman2019spectral}, facilitating learning of high-frequency details in neural fields. 
Further, \citet{mildenhall2020nerf} proposed positional encoding, a special case of Fourier feature encoding, can greatly boosts reconstruction quality for novel view synthesis.
\citet{barron2021mipnerf}~extended positional encoding to \textit{integrated} positional encoding to enable \textit{anti-aliased} multi-scale neural rendering. 
\citet{hu2023Tri-MipRF}~show the effectiveness of integrated positional encoding when applied to the multi-scale hybrid tri-plane representation. 
These methods necessitate \textit{explicit multi-scale supervision} through downsampling images~\cite{barron2021mipnerf, xi2022bungeenerf} and are primarily designed for filtering in 2D screen space ---- for novel-view synthesis applications. 
Our approach in contrast has broader applicability, as it can be employed on any neural field, and eliminates the need for a downsampled supervision signal.

\paragraph{Frequency decomposition by network design}
Neural fields can decompose the frequency of the signal by specific neural network design. Multiplicative filter networks~(MFNs)~\cite{fathony2021multiplicative} introduce an architecture that outputs a linear combination of sinusoidal bases, enabling explicit frequency control. BACON~\cite{lindell2021bacon} utilizes a multi-layer MFN for signal reconstruction, providing frequency upper-bound estimates at each layer.
Polynomial Neural Fields~\cite{Polynomial} refines this approach with a more complex architecture, facilitating precise frequency sub-band decomposition, and ~\citet{shekarforoush2022residual} enables progressive training of finer level-of-detail via the introduction of skip connections. 
These methods have an inherent limitation: due to their specialized network design, they are not applicable to mainstream \textit{hybrid} neural field representations.
Some other methods perform joint spatial and frequency decomposition by applying Fourier feature encodings on local grid features~\cite{wu2023nffb, dou2023mfld}.
Despite their hybrid representation, similarly to MFNs, these methods rely on \textit{specialized} architectures.
Our approach can be applied atop \textit{any} neural field and \textit{independently} of its underlying architecture, therefore allowing to maintain the advantages of (mainstream) \textit{hybrid} neural fields, while adding the ability to perform frequency decomposition.
\section{Method}
\label{sec:method}
\begin{figure*}
\begin{center}
\includegraphics[width= \textwidth]{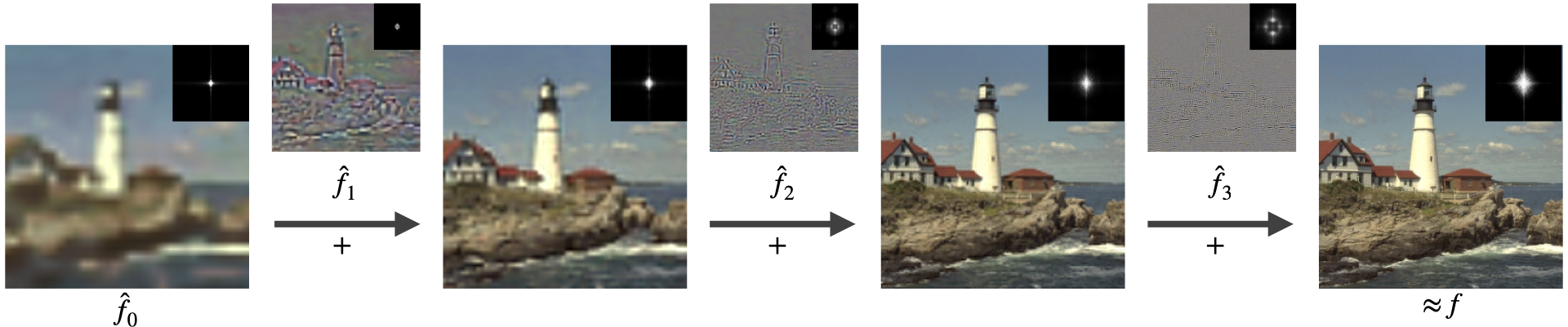}
\end{center}
\vspace{-1em}
\caption{
\textbf{Cascaded training} -- We visualize our neural field decomposition where each level produces a signal with frequency band~$[\freq_{i}, \freq_{i+1}]$.
The decomposition is generated by a cascaded optimization akin to the generation of Laplacian pyramids in signal processing.
Analogously, as the decomposition is quasi-orthogonal, we can generate the signal at any level through superposition. Image taken from the Kodak dataset~\cite{kodak}.
\vspace{-1.2em}
}
\label{fig:cascaded}
\end{figure*}

%
Neural fields $\f(\x; \params)$ represent signals with neural networks, by randomly sampling field locations $\x$, and optimizing the parameters $\params$ to reproduce a given ground truth value $\field(\x)$:
\begin{equation}
    \params^* = \argmin_{\params} \: \expectation_\x
    \| \f(\x; \params) - \f(\x) \|_2^2
    \label{eq:neuralfield}
\end{equation}
Thanks to the commutativity of the Fourier transform and linear operators~\cite{SignalProcessing}, we can represent \textit{any} signal, as the superposition of band-limited signals:
\begin{equation}
    \f(\x) = \f_0(\x) + \dots + \f_K(\x) 
\end{equation}
where the discrete set of frequencies is denoted as $\{\omega_k\}$, and $\omega_0{=}0$, $\omega_{K+1}{=}\infty$, and $\f_k(\x) {=} \f_{[\omega_k, \omega_{k+1}]}(\x)$ is the band-pass filtered $\f(\x)$, with frequency band $[\omega_k, \omega_{k+1}]$.
In our work, we seek to train a neural field with the same characteristics, so that the field decomposes as: 
\begin{equation}
    \f(\x; \{\params_k\}) = \f_0(\x; \params_0) + \dots + \f_{K}(\x; \params_{K}).
    \label{eq:fielddecomp}
\end{equation}
Note that $\params_k$ \textit{only} represents the fraction of the signal with spectra in the range $[\omega_k, \omega_{k+1}]$.
Further, we will \textit{not} assume a filtered version of the ground truth signal $\f_k(\x)$ to be readily available, as this is an unreasonable request for inverse problems; e.g.~we cannot ask for a 3D representation to be filtered, as we only have 2D information about the 3D scene at hand.
In theory, this could be achieved if a filtered version of the signal could be derived by convolving with a low-pass filter $\kernel_{\omega_k}$ with cut-off frequency $\omega_k$:
\begin{equation}
\f_{[0, \omega_k]}(\x) = \kernel_{\omega_k}(\x) \circledast \f(\x) 
\label{eq:lowpass}
\end{equation}
from which we can obtain the desired \textit{band-pass} signal:
\footnote{Note the above resembles the filters employed to construct Laplacian pyramids via repeated applications of the so-called ``mexican hat'' kernel~$\kernel_{\omega_{k+1}}(\x) - \kernel_{\omega_k}(\x)$~\cite{laplacianpyramid}.}
\begin{equation}
\f_k(\x) = \kernel_{\omega_{k+1}}(\x) \circledast \f(\x) -  \kernel_{\omega_k}(\x) \circledast \f(\x) 
\label{eq:bandpass}
\end{equation}

In what follows, we will first show that \eq{lowpass} can be achieved via \textit{optimization} rather than explicit convolution~(\cref{sec:lowpass}), and how~\eq{bandpass} leads to a convenient cascaded training scheme~(\cref{sec:cascaded}).

\subsection{Filtering via optimization}
\label{sec:lowpass}
Denote with $f[t]$ a discrete signal with uniform sampling of period $T$. From classical signal processing~\cite{SignalProcessing}, such a discrete signal can be \textit{reconstructed} as a continuous signal by applying a discrete-continuous convolution with a reconstruction kernel $\hat{\kernel}(\x)$:
\begin{equation}
    \hat{\f}(\x) = \hat{\kernel}(\x) \circledast \sum_t \f[t] \cdot \delta(\x-\x_t)
\label{eq:recon}
\end{equation}
where $\{\x_t\}$ are the sampling locations. Via the Convolution Theorem, it can be shown that $\hat{f}(\x)$ has frequency that is upper-bounded by~$\omega \leq \frac{2\pi}{T}$, and use 
notation $\hat{\f}_\omega(\x)$ to indicate this upper bound.

\begin{figure}[h]
    \centering
    \begin{minipage}[t]{0.17\textwidth}
        \centering
        \vspace{-3.5em}
        \includegraphics[width=\linewidth]{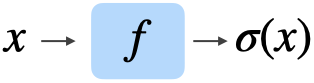} \\
        \vspace{1.3em}
        Neural Field
    \end{minipage}\hfill
    \begin{minipage}[t]{0.28\textwidth}
        \centering
        \includegraphics[width=\linewidth]{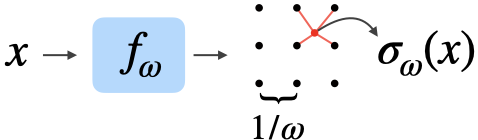} \\
        Band-Limited Neural Field
    \end{minipage}
    \caption{A standard neural field (left). Our band-limited neural field (right), which samples the outputs of a neural field $f_w$ on a regular grid and interpolates to output a \textit{band-limited} signal $\sigma_w(x)$. This enables direct control of the frequency bandwidth $\omega$ of the signal $\sigma_\omega(x)$.}
    \label{fig:pipeline}
    \vspace{-1em}
\end{figure}

We now replace the discrete signal $\f[t]$ in \eq{recon} with a neural field sampled on a uniform lattice~$\{\x_t\}$:
\begin{equation}
    \hat\f_\omega(\x; \params)
    = \kernel_{\omega}(\x) \circledast \sum_t \f(\x_t; \params) \cdot \delta(\x-\x_t)
\label{eq:recon_2}
\end{equation}
We note that when $\kernel_{\omega}(\x)$ is a triangle kernel $\Lambda_\omega(\x)$ with local support equal to the period $T$, then the convolution above amounts to typical linear interpolation that is used to propagate features within voxels~\cite{takikawa2021nglod} that is found in many NeRF implementations~\cite{mueller2022instant}.:
\begin{align}
    \hat\f_\omega(\x; \params) &=
    \Lambda_\omega(\x) \circledast \sum_t \f(\x_t; \params) \cdot \delta(\x-\x_t)
    \\
    &\equiv \text{Interp}(\x; \{\f(\x_t; \params)\})
    \label{eq:interp}
\end{align}
Let us now incorporate~\eq{interp} into the field training loop~\eq{neuralfield}:
\begin{align}
    \params^* &= \argmin_{\params} \: \expectation_\x
    \| \hat\f_\omega(\x; \params) - \f(\x) \|_2^2
\label{eq:filteropt}
\end{align}
Under certain conditions~(\cref{sec:proof}), we can show that:
\begin{equation}
\hat\f_\omega(\x; \params^*) \approx \kernel_{\omega}(\x) \circledast \f(\x) 
\label{eq:equivalence}
\end{equation}
which shows that an optimized neural field $\hat\f_\omega(\x; \params^*)$ can approximate a low-pass filtered signal $\f(\x)$ with frequency band $[0, \omega]$; see \Cref{fig:pipeline}.

\begin{figure}
\begin{center}
\resizebox{\linewidth}{!}{
\setlength{\tabcolsep}{5pt}
\begin{tabular}{c|ccc|ccc|ccc}
    & \multicolumn{3}{c|}{BANF-Linear}& \multicolumn{3}{c|}{BANF-Sinc} & \multicolumn{3}{|c}{BACON} \\
    GT Filter & $1/4\times$ & $1/2\times$ & $1\times$ & $1/4\times$ & $1/2\times$ & $1\times$ & $1/4\times$ & $1/2\times$ & $1\times$ \\ \hline
    Sinc & 28.17 & 29.52 & \multirow{2}{*}{39.55} & 34.68 & 36.23 & \multirow{2}{*}{39.18} & 31.18 & 33.14 & \multirow{2}{*}{38.87} \\
    Linear & 28.61 & 28.82 &   & 30.99 & 32.19 &   & 24.45 & 27.93 &  
\end{tabular}
}
\end{center}

\vspace{-1em}
\includegraphics[width=\columnwidth]{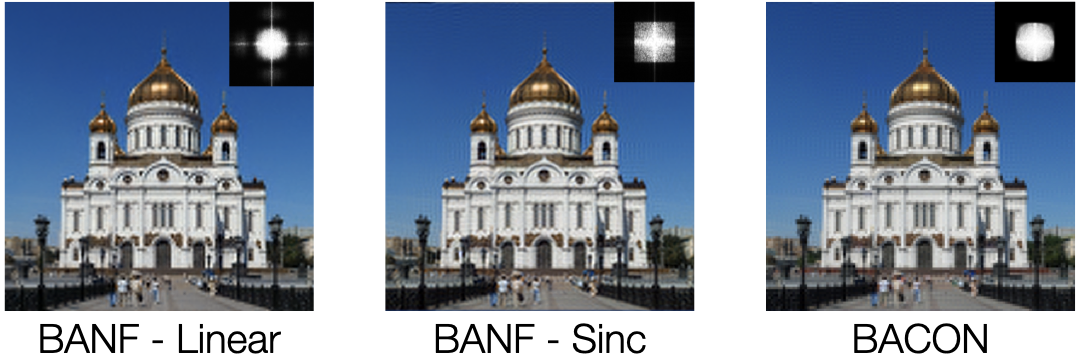}

\captionof{figure}{
\textbf{Interpolation Kernels} -- 
By changing the interpolation kernel we can control the frequency cut in the Fourier domain.
}
\vspace{-1em}
\label{fig:2d_interpolators}
\end{figure}

We note that the choice of kernel $\kernel_{\omega}(\x)$ gives us direct control over how a neural field behaves in the Fourier domain. Depending on the interpolation kernel chosen, we can opt for fast but less precise reconstruction using a \textit{linear} kernel or slower yet highly accurate \textit{sinc} interpolation; see \Cref{fig:2d_interpolators}.

\subsection{Cascaded training} 
\label{sec:cascaded}

Recall our objective is to obtain a decomposition of the field like~\eq{fielddecomp}. With~\eq{filteropt}, we can optimize our first neural field $\hat\f_0(\x; \params_0)$ with frequency band $[0, \omega_1]$:
\begin{align}
    \params^*_0 &= \argmin_{\params_0} \: \expectation_\x
    \| \hat\f_0(\x; \params_0) - \f(\x) \|_2^2
\end{align}

We proceed with our next neural field  $\hat\f_1(\x; \params_1)$ with frequency band $[\omega_1, \omega_2]$:
\begin{equation}
\params^*_1 = \argmin_{\params_1} \: \expectation_\x
\| \hat\f_1(\x; \params_1) - (\f(\x)-\hat\f_0(\x; \params_0^*)) \|_2^2
\label{eq:windowed}
\end{equation}
We note that $(\f(\x)-\hat\f_0(\x; \params_0^*))$ is a high-pass filtered signal with frequency band $[\omega_1, +\infty]$, and as we have shown in ~\eq{equivalence}, $\hat\f_1(\x; \params_1)$ will be a low-pass signal with frequency band $[0, \omega_2]$.
Therefore, the optimization above will capture a representation whose frequency band $[\omega_1, \omega_2]$ is the intersection of these two ranges.
The optimization scheme for subsequent levels can then be derived by induction:
\begin{equation}
\argmin_{\params_k} \: \expectation_\x
\| \hat\f_k(\x; \params_k) - (\f(\x)-\Sigma_{l=0}^{k-1} \hat\f_l(\x; \params_l^*)) \|_2^2
\label{eq:alllevels}
\end{equation}
Once all levels are trained, the signal can then be recomposed via superposition of frequency bands:
\begin{align}
\hat{\f}(\x; \params^*) &= \sum_k \hat\f_k(\x; \params^*_k) \approx \f(\x)
\end{align}
This straightforward, yet powerful, idea can be applied to a variety of signal reconstruction tasks.

\paragraph{Implementation}
To implement cascaded training, unless specified otherwise, we initialize all neural field parameters of all levels with random zero-mean weights of near-zero variance and zero bias.
Further, the lowest resolution model~$\hat{\f}_0(\x; \params_0)$ is trained gradually; following~\cite{yang2023freenerf, hertz2021sape}, we initially train the model in a warm-up phase at $1/4 \times$ and then $1/2 \times$ of its intended final resolution.
We empirically found that having this warm-up phase encourages smoothness and stabilizes optimization.
We ablate our choice in supplementary 

\section{Experiments}
\label{sec:results}
We demonstrate the efficacy of our method in frequency decomposition with three distinct applications:
2D image fitting~(\cref{sec:images}), 3D shape fitting with signed distance field supervision~(\cref{sec:surface_fitting}), and 3D shape recovery from inverse rendering~(\cref{sec:neus}).
We then ablate our design choices in ~(\cref{sec:ablation}). We utilize the linear interpolation kernel in all experiments unless otherwise specified.

\begin{figure}
\begin{center}
\includegraphics[width=.9\columnwidth]{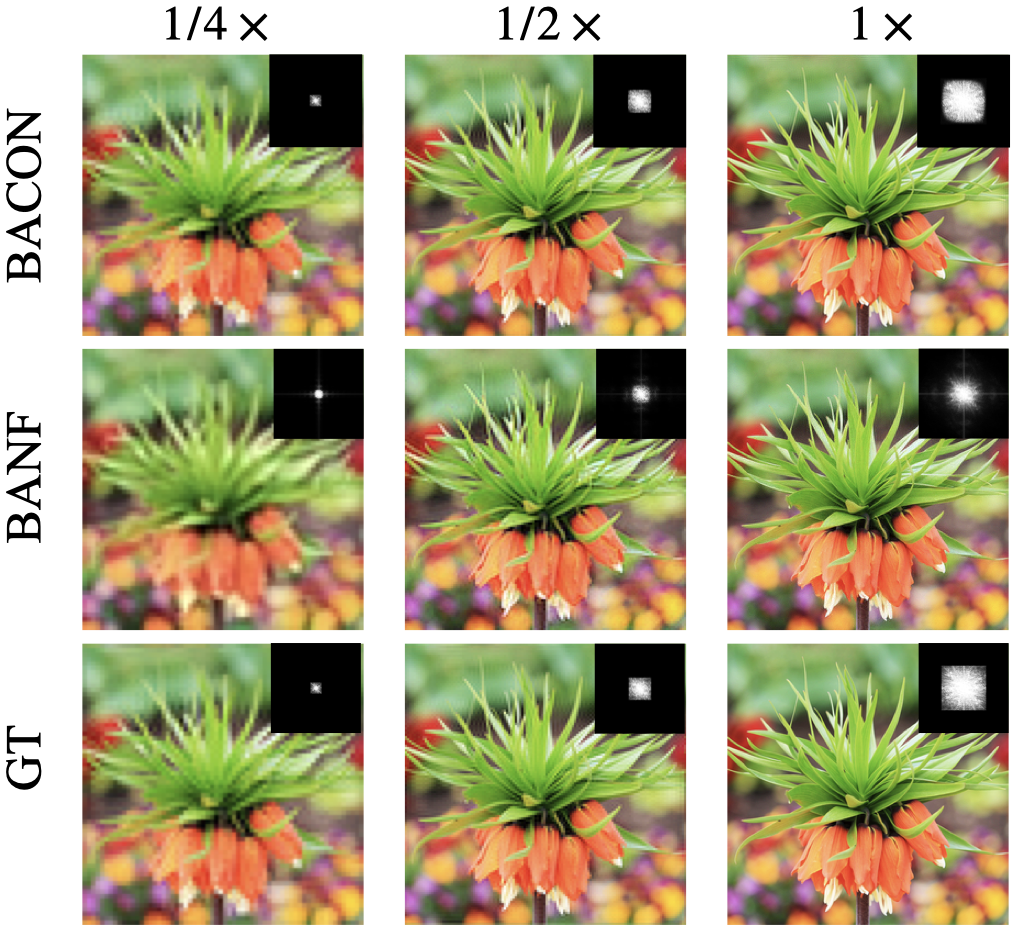}
\end{center}
\vspace{-1em}
\caption{
\textbf{Image fitting and filtering} --
We visualize both multi-scale reconstruction and Fourier spectra of BANF compared against BACON and the ground truth image.
}
\vspace{-1.5em}
\label{fig:2d}
\end{figure}

\subsection{Image filtering (2D)}
\label{sec:images}
We evaluate our method qualitatively by fitting a neural field to 2D images and reconstructing each image at four different frequency levels.
We use an iNGP~\cite{mueller2022instant} backbone for images, with an MLP with~$3$ hidden layers, each with~$32$ neurons to convert deep features into color.
The ground truth signal is a~2D image of resolution $256^2$, for which we learn to represent it at resolution $r \in \{64^2,128^2,256^2\}$. 
We train the network with RMSProp~\cite{rmsprop} for 1K iterations with a learning rate of $2{\times}10^{-3}$.
We use a batch size of $2^{16}$ randomly sampled points.
We note that when sampling, we sample from a continuous 2D space with bilinear interpolation, as we are interested in the behavior of each method related to the underlying \emph{continuous} signal.
We use the mean squared error as the training objective.

We evaluate our method qualitatively on images from the DIV2K dataset~\cite{div2k}, resampled to a resolution of $256^2$ following~\cite{lindell2021bacon}. 
We select BACON~\cite{lindell2021bacon} as a representative baseline in 2D image decomposition using the Multiplicative Filter Network~(MFN)~\cite{fathony2021multiplicative} architecture.
As a reference, we provide images decomposed directly with classical signal processing, which we refer to as ground truth.

As shown in \Cref{fig:2d}, our method successfully approximates low-pass filtering, being similar to the one provided by BACON as well as the ground truth.
We note that as we use linear interpolation, we do not obtain exact low-pass filtering, but rather an approximation~(The Fourier transform of a triangle kernel is $sinc^2$, which is only an approximate low-pass filter).
Yet, as shown, it approximates well and leads to a meaningful frequency decomposition of the signal.
Also, note that there is visible Gibbs ringing in both BACON and the ground truth at the lower scales. Our results, on the other hand, while faithfully representing the details at lower frequency, do not suffer from these artifacts.

In \Cref{fig:2d_interpolators}, we demonstrate the adaptability of our method to various interpolation kernels. We assess our method's performance using \textit{sinc} (of order 6 with box window) and \textit{linear} interpolation kernels with circular interpolation support, resulting in sharp frequency cut and alias-free results respectively. Additionally, we quantitatively compare the reconstructed output and ground truth filtered using either \textit{linear} and \textit{sinc} filters. 

\begin{figure}
\begin{center}
\includegraphics[width=\columnwidth]{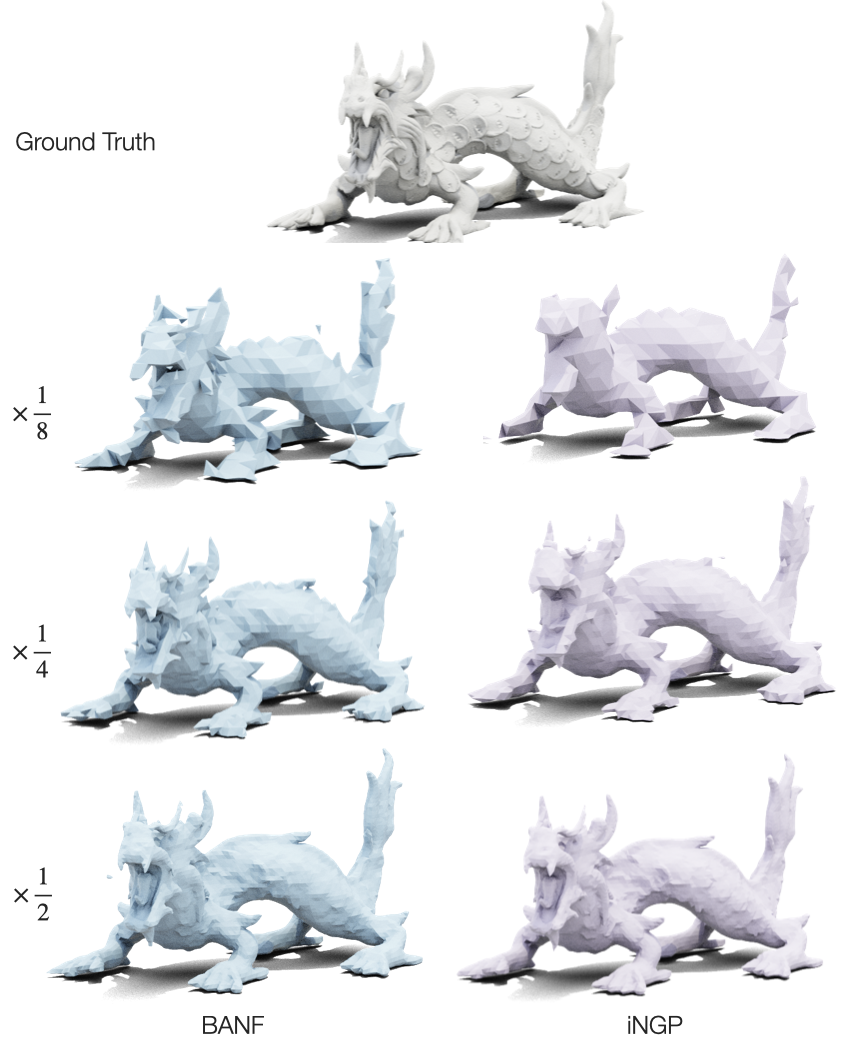}
\vspace{.5em}
\resizebox{\linewidth}{!}{
\setlength{\tabcolsep}{12pt}
\begin{tabular}{c|ccc|ccc}

                        & \multicolumn{3}{c}{Asian Dragon}                           & \multicolumn{3}{c}{Thai Statue} \\ 
method                        & 32                        & \multicolumn{1}{c}{64} & 128   & 32        & 64       & 128      \\ \hline
iNGP                    & \multicolumn{1}{r}{1.97}  & 0.79                  & \textbf{0.32} & 1.83     & 0.82    & \textbf{0.38}    \\
BANF                    & \multicolumn{1}{r}{\textbf{1.23}} & \textbf{0.65}  & \textbf{0.32} & \textbf{1.41}     & \textbf{0.68}    & \textbf{0.40}   
\end{tabular}
}
\end{center}
\vspace{-2em}
\captionof{figure}{
\textbf{Fitting SDFs} -- 
We show qualitative results and report the Chamfer-L2 distance ($\times 10^2$) $\downarrow$.
By filtering the signal during training, our method implements anti-aliasing, so that the extracted signal at coarser scales approximates better. 
}
\vspace{-1.5em}
\label{fig:3d-sdf}
\end{figure}

\subsection{Signed Distance Fields (3D)} 
\label{sec:surface_fitting}
We now evaluate our method for learning Signed Distance Fields~(SDF) at various levels of detail through \textit{direct} 3D supervision.
We assume that exact meshes are unavailable, and only implicit signed distances are, which is commonly the case for real-world captures.
We demonstrate that our method can directly decompose frequencies.

\paragraph{Dataset}
We evaluate with two commonly used meshes ``Asian Dragon'' and ``Thai Statue''.
Both have intricate details that are of high frequency, making the SDF fitting task and, more specifically frequency decomposition, challenging.
We first normalize the meshes to fit into a unit sphere.
We then sample 500k points to get their respective SDF values, with 40\% of the points on the surface, 40\% near the surface, and the rest uniformly within the bounding box.
We use the same points to train all models, and optimize via mean square error for the predicted SDF values.

\paragraph{Baseline}
We compare our method against iNGP.
We use a multi-resolution hash grid of resolutions 16 to 2048, for 100k iterations with a batch size of 10k points until full convergence.
We then extract the meshes at the same resolutions as before using marching cubes~\cite{MarchingCubes}.
We evaluate each method both qualitatively and quantitatively via Chamfer-L2 distance.

\paragraph{Implementation}
We use the Adam optimizer~\cite{Adam} with a learning rate of $10^{-3}$ and the default training settings provided in iNGP. 
We learn the signal at four different resolutions $r \in \{32^3, 64^3, 128^3, 256^3\}$.
As in~\citet{wang2021neus}, the coarsest level's MLP is specifically initialized to produce a sphere for stable training.
We train with a batch size of 100k points. 
We train each level for 10k iterations, except for the coarsest level, which we found to converge already at 5k iterations.

\paragraph{Discussions}
In~\Cref{fig:3d-sdf}, we show quantitatively and qualitatively that our method better respects the Nyquist sampling theorem~\cite{SignalProcessing} due to the filtering that our method performs, resulting in higher quality approximations at coarse resolutions.
As the resolution increases, the marching cubes sampling frequency exceeds the Nyquist rate, thus, as expected, both our method and vanilla iNGP achieve similar reconstruction quality.


\begin{figure*}
\begin{center}
\includegraphics[width=\textwidth]{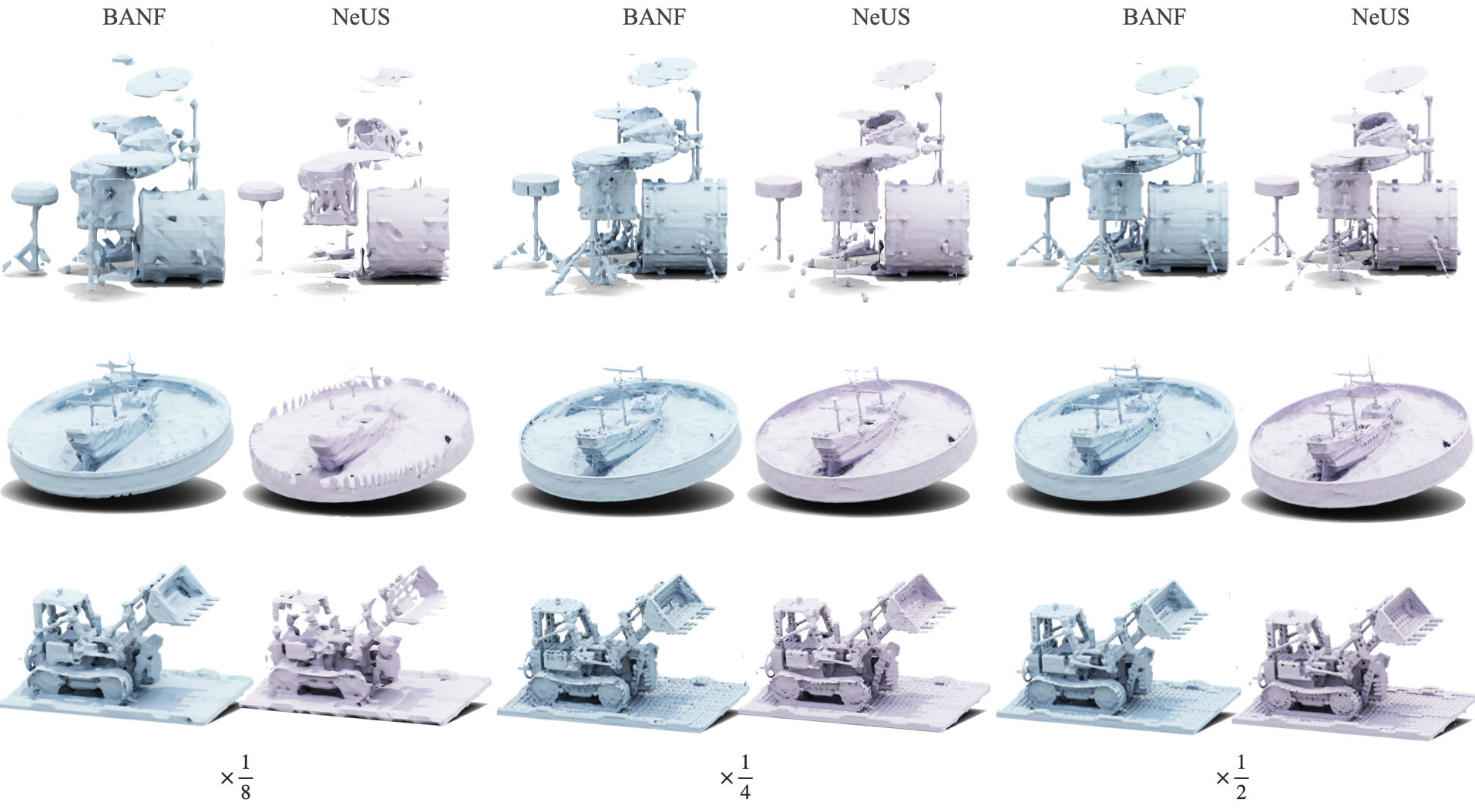}
\end{center}
\vspace{-2em}
\captionof{figure}{
\textbf{Inverse rendering -- }
We integrate our technique into NeUS~\cite{wang2021neus} for level-of-detail surface reconstruction.
We show qualitative results on on the Synthetic NeRF Dataset~\cite{barron2021mipnerf}.
Note how, \textit{especially} at coarser scales, the baseline lacks approximation of fine-grained details~(caused by aliasing, as the baseline lacks filtering).
Quantitative results are provided in \Cref{tab:3d}.}
\label{fig:3d}
\end{figure*}
\begin{table*}[!tb]
\begin{center}
\resizebox{\linewidth}{!}{
\setlength{\tabcolsep}{16pt}
\begin{tabular}{cc|cccccccc|c}
 Scale & Method &Chair & Drums & Ficus & Hotdog & Lego & Material & Mic & Ship & Average \\
 \midrule
\multirow{2}{*}{1/8x} & NeUS & \textbf{6.05} & 21.8 & 40.3 & \textbf{8.82} & 10.2 & 24.5 & 10.5 & 16.4 & 17.3 \\
 & BANF & 7.79 & \textbf{14.4} & \textbf{11.1} & 15.0 & \textbf{9.36} & \textbf{11.6} & \textbf{10.2} & \textbf{11.6} & \textbf{11.4} \\
\midrule
\multirow{2}{*}{1/4x} & NeUS & \textbf{4.09} & 11.41 & 18.9 & \textbf{7.54} & \textbf{5.89} & 17.0 & \textbf{5.60} & 13.2 & 10.5 \\
 & BANF & 6.29 & \textbf{8.66} & \textbf{7.16} & 10.2 & 6.92 & \textbf{9.44} & 7.53 & \textbf{9.32} & \textbf{8.19} \\
\midrule
\multirow{2}{*}{1/2x} & NeUS & \textbf{3.94} & 9.03 & 8.06 & \textbf{7.17} & \textbf{4.74} & 14.7 & \textbf{4.32} & 12.1 & 8.01 \\
 & BANF & 5.92 & \textbf{8.43} & \textbf{5.78} & 10.2 & 6.89 & \textbf{8.90} & 6.66 & \textbf{9.15} & \textbf{7.74} \\
\midrule
\multirow{2}{*}{1x} & NeUS & \textbf{4.08} & 8.45 & 5.75 & \textbf{7.12} & \textbf{4.48} & 13.7 & \textbf{4.30} & 11.8 & \textbf{7.45} \\
 & BANF & 5.82 & \textbf{8.31} & \textbf{5.36} & 10.0 & 6.69 & \textbf{8.68} & 5.79 & \textbf{8.95} & \textbf{7.45} \\
\end{tabular}
}

\label{tab:3d-blender}
\end{center}
\vspace{-1em}
\captionof{table}{
\textbf{Inverse rendering} --
Quantitative results for NeUS~\cite{wang2021neus} LOD reconstruction on the Synthetic NeRF Dataset~\cite{barron2021mipnerf}. See qualitative results in \Cref{fig:3d}. Metric is Chamfer Distance ($\times 10^{-2}) $~$\downarrow$.
\vspace{-1em}
}
\label{tab:3d}
\end{table*}
\begin{table*}[]
\centering
\resizebox{\linewidth}{!}{
\setlength{\tabcolsep}{10pt}
\begin{tabular}{cc|cccccccccccc|c}

Scale & Method& Ben & Boat & Bridge & Cabin & Camera & Castle & Colos & Jeep & Bus & Motor & Satel & Shuttle & Avg. \\
 \midrule
\multirow{2}{*}{1/4x} & NeUS & 20.4 & 59.5 & 9.25 & 5.52 & 5.40 & 26.6 & 15.5 & 3.67 & 10.0 & 3.54 & 7.08 & 19.3 & 15.5 \\
 & BANF & \textbf{10.0} & \textbf{6.40} & \textbf{4.61} & \textbf{4.64} & \textbf{2.78} & \textbf{17.4} & \textbf{12.9} & \textbf{2.96} & \textbf{5.48} & \textbf{2.73} & \textbf{5.27} & \textbf{13.5} & \textbf{7.39} \\
\midrule
\multirow{2}{*}{1/2x} & NeUS & 8.04 & 6.15 & 5.01 & \textbf{3.95} & 3.94 & 11.4 & \textbf{9.90} & 2.60 & 5.66 & \textbf{2.15} & \textbf{4.17} & 6.78 & 5.81 \\
 & BANF & \textbf{7.13} & \textbf{4.21} & \textbf{4.63} & 4.42 & \textbf{2.76} & \textbf{10.7} & 10.0 & \textbf{2.58} & \textbf{5.18} & 2.37 & 4.56 & \textbf{5.93} & \textbf{5.38} \\
\end{tabular}
}
\caption{\textbf{Inverse rendering -- } We show qualitative results on the MobileBrick dataset~\cite{mobilebrick}. Metric is Chamfer Distance ($\times 10^{-3}) $~$\downarrow$.
Again, larger gains are observable at coarser scales, when signals are sampled at a frequency that is below the one required by Nyquist.
}
\vspace{-1em}
\label{tab:3d-mobilebrick}
\end{table*}

\subsection{Inverse rendering (3D from 2D)} 
\label{sec:neus}
We demonstrate the applicability of our method in a more complex scenario where only 2D images of a 3D signal are available. 
Here, we tackle the inverse problem of 3D surface reconstruction at \textit{multiple} levels of detail. 
Note that as a uniformly sampled 3D signal is not available (only 2D observations are available) Fourier analysis cannot be used to create supervision.

\paragraph{Dataset and baseline} 
For the baseline we use vanilla NeUS~\cite{wang2021neus} followed by marching cubes~\cite{MarchingCubes} to extract meshes. 
We evaluate our method on the~(Synthetic) NeRF dataset~\cite{barron2021mipnerf} and the~(real) MobileBrick dataset~\cite{mobilebrick}.
For MobileBrick, for both our method and NeUS, we use mask supervision and train without background modeling. 
We evaluate each method both qualitative, and quantitatively via the \mbox{Chamfer-L2} distance.

\paragraph{Implementation}
We extend the surface reconstruction baseline provided by NeUS~\cite{wang2021neus} with our method. 
Following NeUS, we optimize the 3D surface field using volume rendering, guided by supervision from 2D images.
Specifically, given a 3D position $\x {\in} \mathbb{R}^3$ and a viewing direction~$\view {\in} \mathbb{S}^2$, we train neural fields for both signed distance~$\hat{\sdf}(\x; \params)$ and color~$\hat{c}(\x, \view; \params)$. 
We apply cascaded training~(\cref{sec:cascaded}) only to the SDF MLP and directly train separate color MLPs at each level -- this is due to the challenge in color MLP optimization with signal decomposition~(details in \cref{sec:ablation}). 
We use an iNGP~\cite{mueller2022instant} backbone with an MLP that predicts SDF values and a feature vector.
The feature vector is then fed into the color MLP to output RGB values.
We supervise training via classical photometric reconstruction error, along with Eikonal and Laplacian regularizers. 
When implementing our method, we observe a performance degradation when the color MLP relies solely on the output feature vector of the decomposed SDF MLP~(see \cref{sec:ablation} for details).
We thus additionally concatenate the features from the hashgrid to the inputs of the color MLP.
We note that this minor architectural change does not lead to any significant changes in the performance of the baseline NeUS.

As done in ~\cref{sec:surface_fitting}, the coarsest SDF MLP head is initialized to produce a sphere. 
For initializing the color MLP heads, we first initialize the coarsest level randomly without any special treatment.
For the subsequent levels, we bootstrap their training by initializing them with the previous level's~(converged) weights.
Our models are trained at four different resolutions $r \in \{64^3,128^3,256^3,512^3\}$, trained in cascade for 20k iterations, with a batch size of 5k rays.
To compute gradients of SDF that are used both as an input for the color MLP, and for computing the Eikonal and Laplacian regularizers, we use finite differences with the delta value set to ${1}/{(4\times r)}$ for each level of our method, and ${1}/{1024}$ for the baseline model.

\paragraph{Discussions}
As shown in \Cref{tab:3d,tab:3d-mobilebrick}, our method significantly outperforms NeUS on average.
In \Cref{fig:3d}, we visually see improvements on thin structures (see drum stool legs, ship's mast, and loader's bucket).
We further observe improvement in anti-aliasing on the ship's rim.

\subsection{Ablations} 
\label{sec:ablation}

\paragraph{Other feature grid representations.}
Our method is agnostic to the underlying representation.
We test 3D reconstruction with grid and MLP-backed representations on 4 objects from Stanford dataset (we follow BACON, and use their preprocessed data and sampling scheme).
Note how the representation \textit{does not} heavily influence reconstruction quality~(Chamfer-L2 $\downarrow$ $\cdot 10^{-2}$):
\vspace{-.25em}
\begin{center}
\resizebox{\linewidth}{!}{
\begin{tabular}{ccc|ccc|ccc}
    \multicolumn{3}{c|}{HashGrid} & \multicolumn{3}{|c|}{Dense} & \multicolumn{3}{|c}{MLP} \\
    32 & 64 & 128 & 32 & 64 & 128 & 32 & 64 & 128 \\ \hline
     1.35 & 0.82 & 0.54 & 1.38 & 0.83 & 0.55 & 1.37 & 0.79 & 0.58
\end{tabular}}
\end{center}
\section{Conclusions}
\label{sec:conclusions}
We introduce a way to train neural fields that enables frequency decomposition of the represented signals.
Departing from heuristic- or architecture-based approaches, we realize this by a simple modification to the training process.
Our approach is versatile, accommodating various neural field architectures (both fully neural and hybrid representations), and makes \textit{no assumptions} about the training data, such as the availability of pre-filtered signals.
We demonstrate its applicability across a number of neural fields workloads, and in particular by testing its effectiveness for anti-aliased level-of-detail reconstruction.

\paragraph{Limitations}
In this work, we focused on uniformly sampled signals, and it would be interesting how to extend the method to ``contracted'' representations that are commonly used in NeRF to deal with unbounded signals. Additionally, in our current implementation, the method becomes highly memory-intensive at higher grid resolutions. This issue could potentially be addressed through the adoption of more efficient querying strategies and the CUDA-based implementation.

\appendix
\section{Appendix}

\label{sec:proof}
\newcommand{\F}{\mathbf{F}}
\newcommand{\matA}{\mathbf{A}}
\newcommand{\vecb}{\mathbf{b}}
Without loss of generalization to higher dimensions, let us consider 1D signals.
Let us start by replacing $\expectation_\x$ in \eq{filteropt} with a sum operator, hence reducing our stochastic gradient descent optimization with least square optimization.
Samples~$\{\x_n\}$ are drawn uniformly with a period sufficient to satisfy the Nyquist theorem for the ground truth signal $\f$:
\begin{equation}
\argmin_{\params} \:\: \sum_{\{\x_n\}}
\| \text{Interp}(\x_n; \{\f(\x_t; \params)\}) - \f(\x_n) \|_2^2
\end{equation}
Let us rewrite this expression in matrix form by denoting $\{\f(\x_n)\}$ as $\vecb {\in} \mathbb{R}^{N \times D}$, $\{\f(\x_n; \params)\}$ as $\mathbf{X} {\in} \mathbb{R}^{T \times D}$, and by storing in the \textit{skinny} matrix $\matA {\in} \mathbb{R}^{N \times T}$ the linear interpolation coefficients corresponding to the positions $\{\x_n\}$:
\begin{equation}
\argmin_{\mathbf{X}} \:\: \| \matA \mathbf{X}   - \vecb \|_2^2
\end{equation}
Assuming the matrix $\matA$ is invertible, the optimization above provides the closest point projection of $\vecb$ onto the range of~$\matA$~\cite[Sec.~5-3]{boyd}.
Recalling $\matA$ represents functions whose frequency is upper-bounded by $\omega$, this implies the optimization projects our true signal onto the closest function with bounded spectra, which is equivalent to convolving the signal with a low-pass filter as in~\eq{equivalence}.
Our approximations stem from the fact that \textit{linear} interpolation is only an \textit{approximation} of a low-pass filter (with a $\text{sinc}^2$ frequency spectra), and that our optimization is a stochastic gradient descent rather than a close-form solve.

\paragraph{Acknowledgments}
We would like to thank David Fleet, David Lindell, Shayan Shekarforoush for their valuable feedback. In addition, we are grateful to the BIRS workshop on "generative 3D modeling" that made this discussion possible.
Andrea Tagliasacchi is supported by the NSERC discovery grant [2023-05617], and by the SFU Visual Computing Research Chair.

\clearpage
{
    \small
    \bibliographystyle{ieeenat_fullname}
    \bibliography{main}
}

\clearpage
\maketitlesupplementary
\section{Radiance field decomposition}

\begin{figure}[!tb]
\begin{center}
\includegraphics[width=\columnwidth]{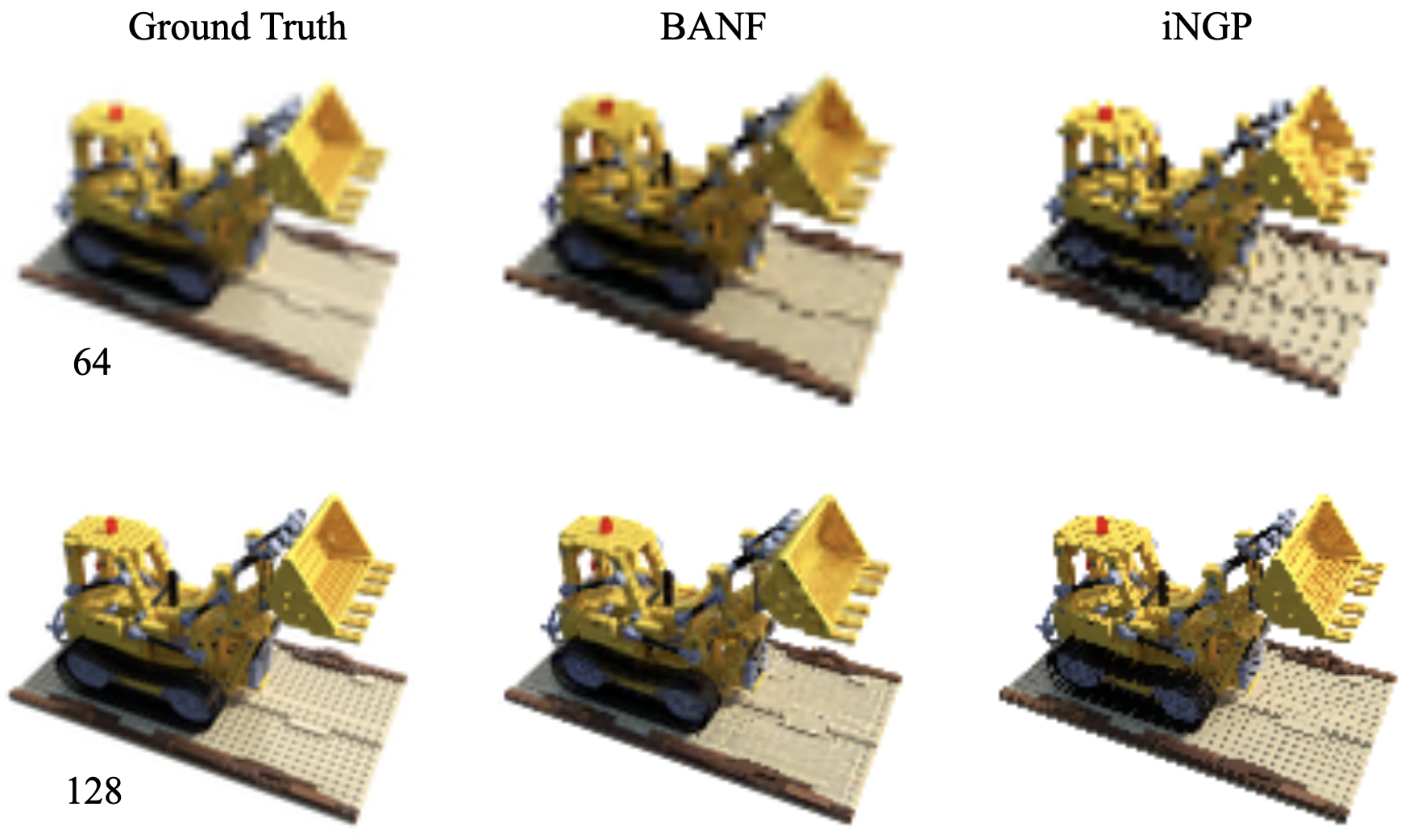}
\vspace{.5em}
\resizebox{\linewidth}{!}{
\setlength{\tabcolsep}{12pt}
\begin{tabular}{ccc|cc}
 Scene    & iNGP-64 & BANF-64 & iNGP-128 & BANF-128 \\ \hline
Chair    &    26.27     &   \textbf{ 30.25}     &   28.89       &  \textbf{32.21}        \\
Drums    &    21.85     &    \textbf{24.33}     &   23.98       &  \textbf{24.86}        \\
Ficus    &    24.06     &    \textbf{27.87}     &   \textbf{27.59}       &  27.45        \\
Hotdog   &    28.33     &    \textbf{30.48}     &   31.52       &  \textbf{32.77}        \\
Lego     &    23.64     &    \textbf{26.59}     &   27.22       &  \textbf{28.12}        \\
Material &    23.35     &    \textbf{24.89}     &   26.30       &  \textbf{26.48}        \\
Mic      &    24.32     &    \textbf{26.77}     &   27.73       &  \textbf{30.36}        \\
Ship     &    22.66     &    \textbf{24.30}     &   24.17       &  \textbf{24.52}        \\ \hline
Average  &    24.31     &   \textbf{26.93}      &   27.17       &  \textbf{28.34}       
\end{tabular}
}
\end{center}
\vspace{-2em}
\captionof{figure}{
\textbf{Color decomposition} -- 
We show an application of our method in frequency decomposition of the color field in NeRFs. 
}
\label{fig:3d_color_supp}
\end{figure}

We extend the application of our frequency decomposition technique to the color field in neural radiance fields. 
Here, we adopt the spherical harmonics (SH) representation proposed in Plenoxels ~\cite{yu2022plenoxels} and~\cite{yu2021plenoctrees} to perform color frequency decomposition. 
Specifically, in addition to implementing frequency-bounded grids for the density field, we apply frequency constraints to the Spherical Harmonics (SH) coefficients. 
At each level of detail, both the density and SH coefficients are queried at a specific resolution and then trilinearly interpolated to determine the density and SH coefficients of the target point.
Subsequently, the computed SH coefficients are transformed into RGB values. 
It's important to note that, due to the linearity of spherical harmonics, constraining the frequency of SH coefficients directly imposes a constraint on the predicted color frequency. 
Further, both the density and color heads are trained using the previously proposed cascaded scheme. 
\quad
We compare the results of this method to a vanilla iNGP queried at target resolutions at test time. 
In~\Cref{fig:3d_color_supp}, we report superior performance to the baseline, highlighting the robustness of our method to aliasing effects.

\paragraph{Implementation} We train iNGP and our variation of it for 50K iterations with a batch size of 4096 rays. 
We evaluate our method on the NeRF Synthetic Dataset~\cite{mildenhall2020nerf} at resolutions $\{64^2, 128^2\}$, while the input images are at $800^2$ resolution. 
These evaluations are performed using grid resolutions $\{32^3, 64^3\}$ respectively, which were empirically determined to yield the best results. 
The SH coefficients of second order were used similar to ~\cite{yu2022plenoxels} and~\cite{yu2021plenoctrees}. 
{Further, we note that we do not employ the extra skip connection from hash grid to the color MLP, as the color is being filtered.}
\vspace{-0.5em}
\section{Image filtering (cont'd)}
\begin{figure*}
\begin{center}
\includegraphics[width=\textwidth]{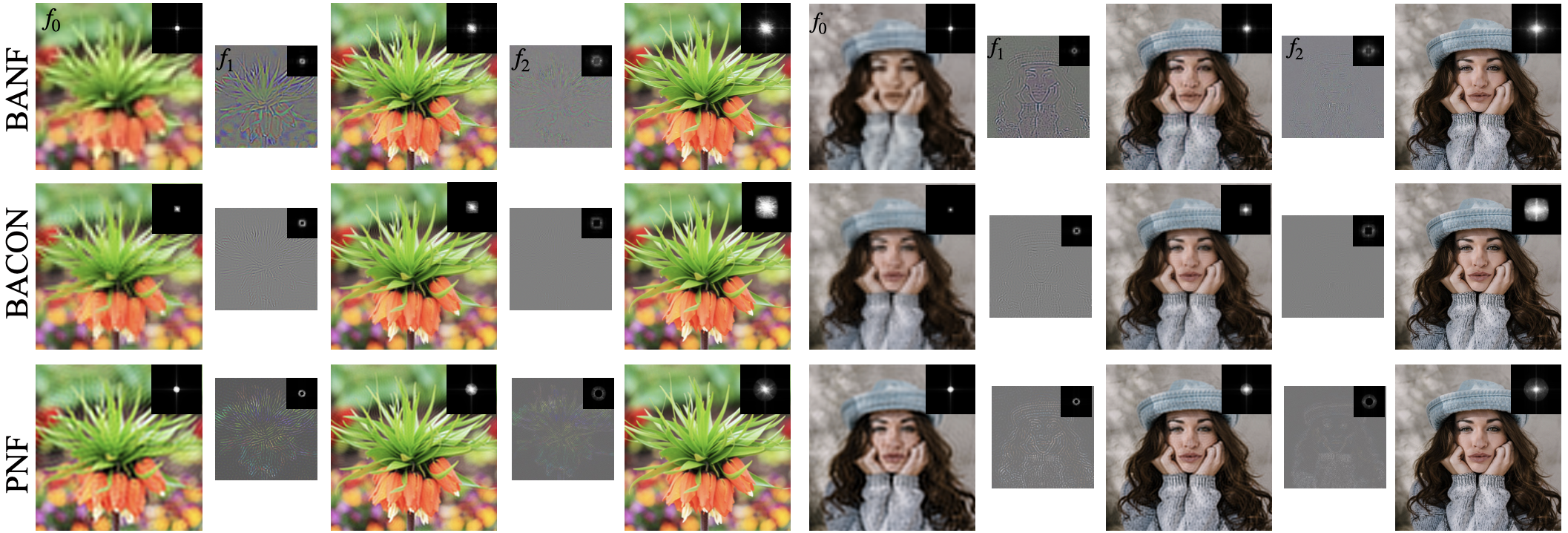}
\includegraphics[width=\textwidth]{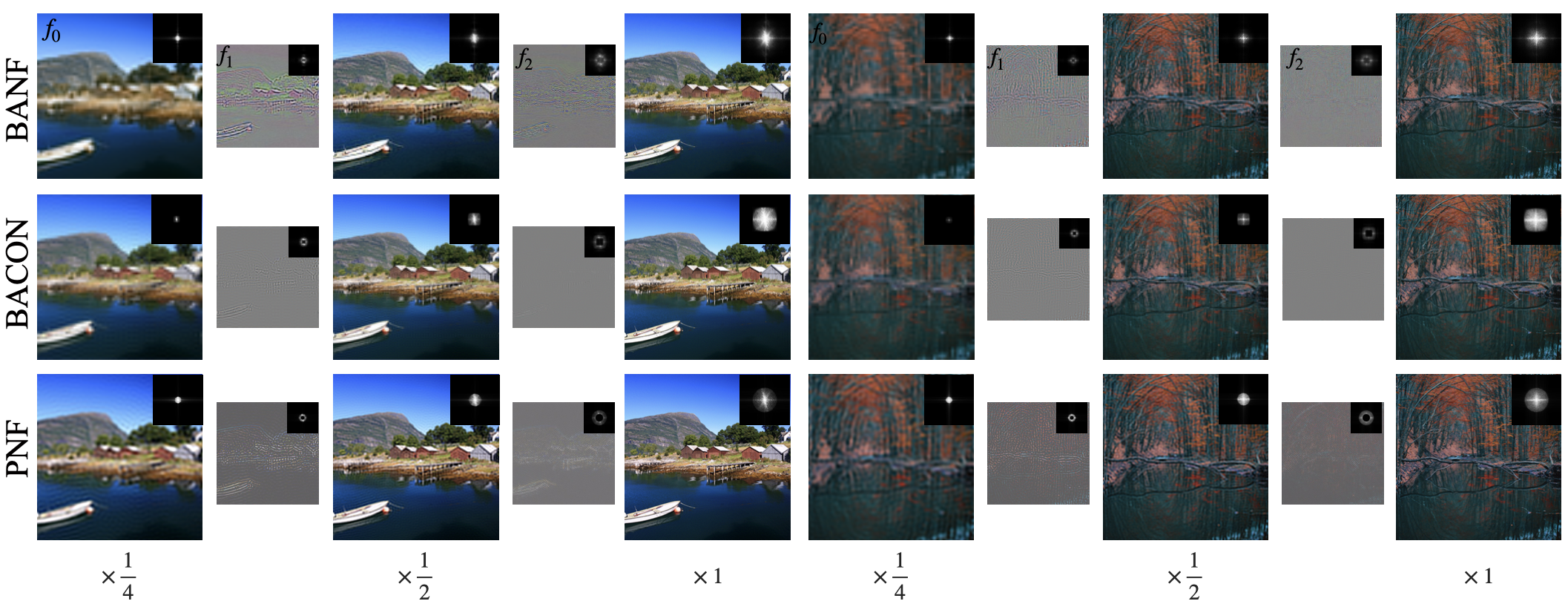}
\end{center}
\vspace{-1em}
\captionof{figure}{
\textbf{Image Filtering -- }
Comparison of 2D filtering results on DIV2K~\cite{div2k} to BACON~\cite{lindell2021bacon} and PNF~\cite{pnf}.
Quantitative results are provided in \Cref{tab:2dsuppl}.
\vspace{1em}
}
\label{fig:2dsupp}
\end{figure*}

We show additional quantitative and qualitative results on the DIV2K~\cite{div2k} dataset in \Cref{fig:2dsupp,tab:2dsuppl}. 
Our method is trained on original images downsampled to $256^2$ resolution and compared to BACON~\cite{lindell2021bacon} and PNF~\cite{pnf}, trained in a similar fashion. 
The networks are trained on 5K samples of the images. 
Evaluation is done at resolution $512^2$, comparing to original images downscaled to this resolution to compute PSNR. Qualitative results are shown at resolutions~$\{64^2, 128^2, 256^2\}$.
As our method is compatible with any neural field, we demonstrate that, being based on an efficient backbone, it can reconstruct high-quality multi-resolution images while maintaining the same number of parameters as other techniques.

\begin{table}[!bt]
    \centering
    \resizebox{.8\linewidth}{!}{
    \begin{tabular}{c|ccc}
        Method        & BACON  & PNF    & BANF            \\ \hline
        PSNR $\uparrow$         & 29.266 & 29.470 & \textbf{30.455} \\
        \# Parameters $\downarrow$ & 0.268M & 0.276M & \textbf{0.244M}
    \end{tabular}
    }
    \vspace{-.5em}
    \caption{\textbf{Image fitting} -- quantitative results on DIV2K~\cite{div2k}.}
    \label{tab:2dsuppl}
\end{table}
\setlength{\tabcolsep}{12pt}

\begin{table}[!tbh]
\centering
\resizebox{.8\linewidth}{!}{
\begin{tabular}{c|ccc}

   Resolution      & 64             & 128            & 256           \\ \hline
Bilinear & 25.650              & 25.364              & 30.455             \\
Bicubic  & \textbf{26.559}     & 26.380              & \textbf{30.697}    \\
Lanczos  & 26.050              & \textbf{26.415}     & 30.210        \vspace{-1em}
\end{tabular}
}
\captionof{table}{PSNR reported on interpolation with higher order kernels.}
\label{tab:int_supp}
\end{table}
\begin{figure*}[h]
    \centering
    \includegraphics[width=\textwidth]{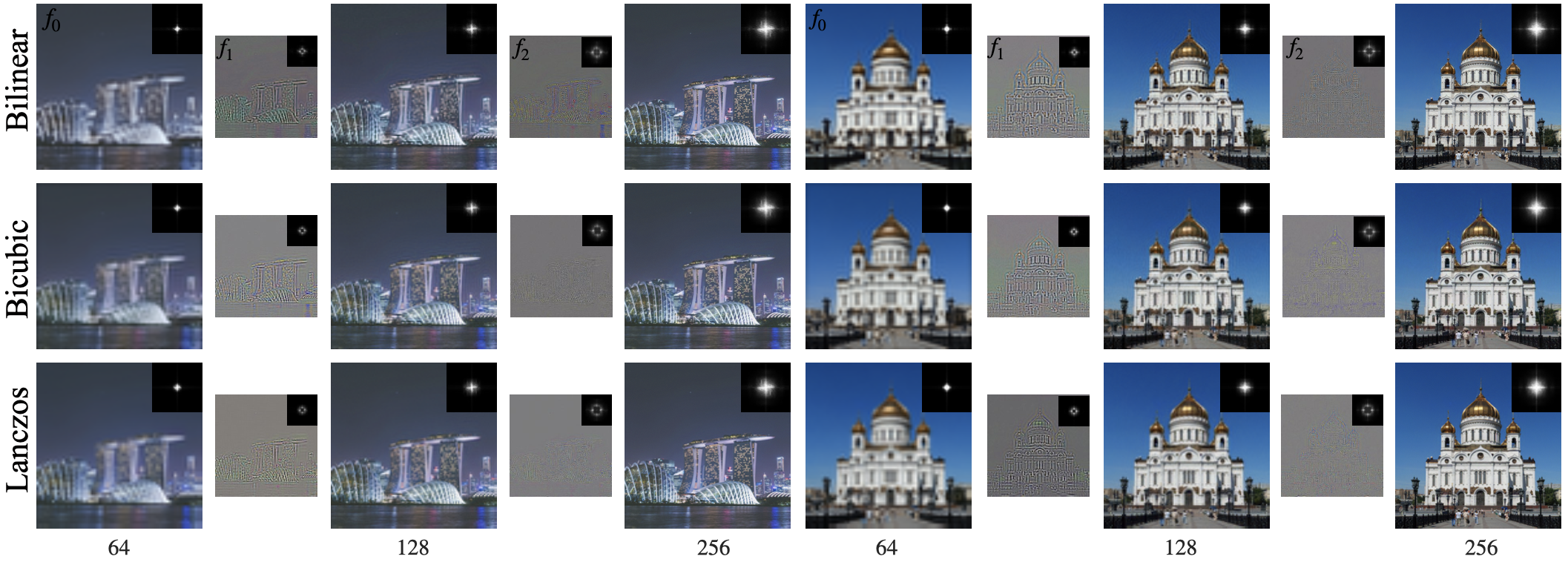}
    \caption{
        \textbf{Higher Order Kernels -- }
        We evaluate the effect of higher order interpolation kernels on DIV2K~\cite{div2k}.
    }
    \label{fig:2dint}
\end{figure*}

\section{Higher order kernels}
In our main results, we used \textit{linear} interpolation in our filtering algorithm.
However, it was noted that this interpolation exhibits leakage, attributed to the $\text{sinc}^2$ Fourier transform of the filter. 
We further investigate how higher-order interpolation kernels such as Lanczos and Bicubic can further improve the filtering quality.
In \Cref{fig:2dint}, we evaluate results on the DIV2K~\cite{div2k}
dataset, with images downsampled to $256^2$. We then evaluate both quantitatively and qualitatively at resolutions~$\{64^2, 128^2, 256^2\}$. 
We show in~\cref{tab:int_supp} that higher-order interpolations can help give an extra boost the performance of our method, while producing (perceptually) better filtered images~\cref{fig:2dint}.

\section{Ablations} 
\label{sec:ablations}

We validate our design choices in terms of Chamfer Distance~($\text{CD}\downarrow$) evaluated on the NeRF Synthetic dataset~\cite{barron2021mipnerf} averaged over all objects, and at $1/8\times$ scale. 

\paragraph{Color MLP decomposition}
When we train the color MLP output in a cascaded manner , performance drops ~$\text{CD}{=}8.19 {\rightarrow} 9.16$. 
We attribute this to the color MLP relying on the SDF MLP output.
As color and geometry exhibit different frequencies in terms of their signal, (\eg, a highly complex shape with solid coloring), this reliance would be detrimental.
For this ablation study we evaluate at $1/4\times$ scale, as the difference is less marked at $1/8\times$ scale.

\paragraph{Color MLP Input}
Removing the hash features input to the color MLP  results in a performance drop from  
$\text{CD}{=}11.4 {\rightarrow} 13.6$. 
Without the hash feature input, the color MLP relies only on the SDF features. At low scales, the SDF features will contain mostly low frequency content. We hypothesize that relying only on band-limited features reduces color reconstruction accuracy, resulting in inaccurate optimization of the the neural fields.

\paragraph{Resolution warmup}
When we remove the resolution warmup from the coarsest level field, the performance drops~$\text{CD}{=}11.4 {\rightarrow} 11.9$.
This is because starting optimization with lower resolution encourages smooth structures early on, leading to an easier optimization landscape as also shown in~\cite{yang2023freenerf, hertz2021sape}.

\section{Additional 3D reconstruction results}

\paragraph{ModelNet10 data}
We extend our evaluation of 3D shape fitting to include five models from the ModelNet10-tables dataset, assessed using the Chamfer-L2 metric ($\downarrow$ $\cdot 10^{-2}$). Specifically, we examined 3D objects with the following identifiers: Table403, Table435, Table463, Table468, and Table479. Notably, our method demonstrates the capability to reconstruct thin structures even in low resolution, a feature not exhibited by the iNGP baseline:
\vspace{-.1em}
\begin{center}
\resizebox{\linewidth}{!}{
\begin{tabular}{ccc|ccc}
    \multicolumn{3}{c|}{BANF} & \multicolumn{3}{|c}{iNGP} \\
    32 & 64 & 128 & 32 & 64 & 128 \\ \hline
    \textbf{2.38} & \textbf{2.08} & \textbf{0.31} & 18.62 & 2.31 & 0.44
\end{tabular}}
\end{center}


\begin{center}
\includegraphics[width=1\linewidth]{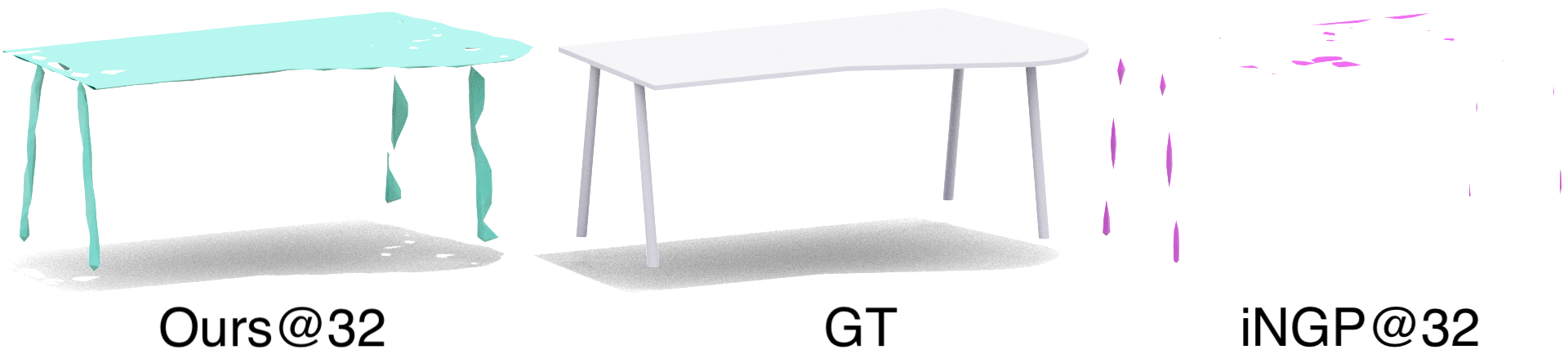}
\end{center}

\paragraph{Additional Baselines for 3D Shape Fitting}
In our comparative analysis, we include BACON \cite{lindell2021bacon} for 3D shape reconstruction. However, we observed that BACON exhibits slower training times and yields lower-quality results, particularly at higher decimation rates, in comparison to BANF. This observation is supported by the Chamfer-L2 metric ($\downarrow$ $\cdot 10^{-2}$) evaluated on the four objects from the Stanford dataset used in BACON. Notably, for this series of experiments, we utilized the BACON codebase for data processing and point sampling pipelines.
\vspace{-.25em}
\begin{center}
\resizebox{\linewidth}{!}{
\begin{tabular}{ccc|ccc|ccc}
    \multicolumn{3}{c|}{BANF} & \multicolumn{3}{|c|}{iNGP} & \multicolumn{3}{|c}{BACON} \\
    32 & 64 & 128 & 32 & 64 & 128 & 32 & 64 & 128 \\ \hline
    \textbf{1.35} & \textbf{0.82} & \textbf{0.54} & 1.98 & 0.91 & 0.61 & 2.65 & 1.00 & 0.58
\end{tabular}}
\end{center}

\paragraph{Qualitative results for MobileBrick dataset}
we also provide a qualitative comparison using the MobileBrick dataset \cite{mobilebrick}. Notably, our method demonstrates superior performance, particularly in handling low-resolution data and thin structures, when compared to NeUS.
\begin{center}
\includegraphics[width=1\linewidth]{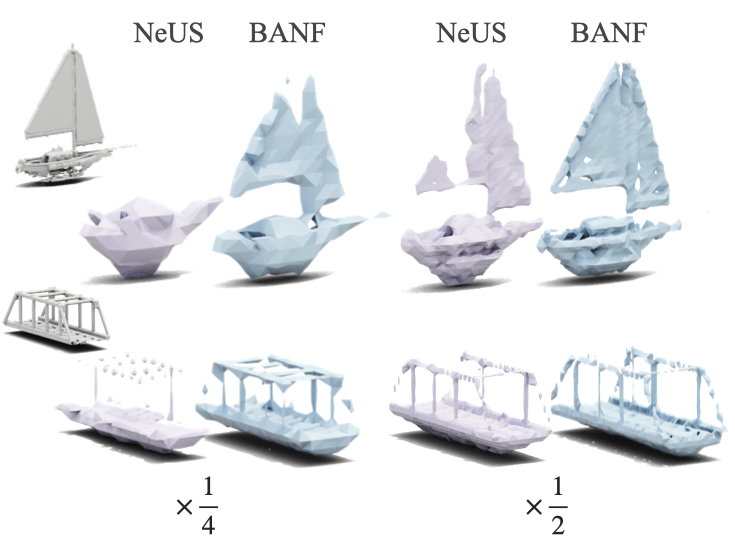}
\end{center}

\section{Evaluation of geometry and color reconstruction on NeRF Synthetic Dataset}

In the NeRF Synthetic Dataset~\cite{mildenhall2020nerf}, not all aspects of the geometries are observable from the training/validation cameras. This includes:

1) Internal structures that evade capture by any means, such as the stem of the plant inside the pot in the "Ficus" scene.

2) Some parts of the geometry that remain invisible from all cameras (both during training and testing). For example, the bottom of the chair in the "Chair" scene.

This inherent limitation results in a skewed assessment of metrics designed to measure the quality of reconstruction. To address this issue when computing the Chamfer Distance, we modify the process for densely sampled points on the surface of a mesh. Specifically, we filter the point cloud so that only points visible from at least one camera are retained. This filtering procedure is applied to both the ground truth and predicted meshes, ensuring a more accurate evaluation.

In inverse rendering, it is expected that the image reconstruction quality degrades as more emphasis is put on (filtered) meshes, since regardless of mesh quality, neural fields can \textit{cheat} to make renderings look good.
Still, this degradation is minimal (${<}1$~dB): average test PSNR for NeuS is 29.31 and 28.45 for BANF.

\section{Proof of filtering via optimization}
In this section, we will briefly review \cref{sec:lowpass} of the paper and then prove that the defined method indeed performs low-pass filtering via optimization.

The parameters of the model are optimized via an L2 loss:

\begin{align}
    \params^* &= \argmin_{\params} \: \expectation_\x
    \| \hat\f_\omega(\x; \params) - \f(\x) \|_2^2
\label{eq:filteropt2}
\end{align}

Using Parseval's theorem \cite{parseval}, we can rewrite the loss function in the Fourier domain (here we are also using linear property of Fourier transform):

\begin{equation}
    \params^* = \argmin_{\params} \: \expectation_\x \| \mathcal{F} \{ \hat\f_\omega(\x; \params) \} - \mathcal{F} \{\f(\x) \} \|_2^2
\label{parseval}
\end{equation}

The forward pass of the model is defined by the following equation (equation \cref{eq:recon_2} from the main paper written in different form):

\begin{equation}
    \hat\f_\omega(\x; \params)
    = \kernel_{\omega}(\x) \circledast [ \f(\x; \params) \cdot \sum_t  \delta(\x-\x_t) ]
\label{eq:forward2}
\end{equation}

Next, we find the Fourier transform of the forward pass of the model to understand the optimization in the Fourier domain (this is basically Discrete-time Fourier transform \cite{dtft}). We use the convolution theorem, which states that multiplication in the spatial domain corresponds to convolution in the frequency domain, and vice versa.

\begin{equation}
    \mathcal{F} \{ \hat\f_\omega(\x; \params) \}
    = \mathcal{F} \{\kernel_{\omega}(\x) \} \cdot [\mathcal{F} \{ \f(\x; \params) \} \circledast  \mathcal{F} \{ \sum_t \delta(\x-\x_t) \} ]
\label{fourier_forward}
\end{equation}

A set of delta functions with period T in the spatial domain transforms to a set of delta functions with period  1/T in the Fourier domain. Furthermore, the spectrum of the kernel $\kernel_{\omega}(\x)$ is a rectangular function (ideally, when using sinc interpolation). Thus, when multiplying a rectangular function with a periodic one, we retain only the central copy of the periodic function:

\begin{equation}
    \mathcal{F} \{ \hat\f_\omega(\x; \params) \}
    = \mathcal{F} \{\kernel_{\omega}(\x) \} \cdot \mathcal{F} \{ \f(\x; \params) \}
\label{fourier_forward}
\end{equation}

Therefore, the loss function becomes:

\begin{equation}
    \params^* = \argmin_{\params} \: \expectation_\x \| \mathcal{F} \{\kernel_{\omega}(\x) \} \cdot \mathcal{F} \{ \f(\x; \params) \} - \mathcal{F} \{\f(\x) \} \|_2^2
\label{final_loss}
\end{equation}

Thus, while optimizing, the loss function pushes the prediction $\mathcal{F} \{ \f(\x; \params) \}$ to match the ground truth $\mathcal{F} \{\f(\x) \}$ within the bounds defined by the rectangular function $\mathcal{F} \{\kernel_{\omega}(\x) \}$ defines. We can control the size and shape of the bound by changing the grid resolution or the type of interpolation kernel. This is clearly illustrated in \cite{graphics_fundamentals_book} chapter 9 figure 9.52

\end{document}